\documentclass[letterpaper, 10 pt, conference]{ieeeconf}  

\IEEEoverridecommandlockouts                              

\usepackage{graphics} 
\usepackage{epsfig} 
\usepackage{times} 
\usepackage{amsmath} 
\usepackage{amssymb}  

\usepackage[T1]{fontenc}
\usepackage{booktabs,multirow,array,caption}

\usepackage{tikz}
\usepackage{textcomp}

\usepackage[pagebackref=true,colorlinks,citecolor=green]{hyperref}
\hypersetup{%
  colorlinks = true,
  linkcolor  = green
}

\newcommand{\R}{\mathbb{R}}

\newcommand\copyrighttext{%
  \footnotesize \textcopyright © 2022 IEEE. Personal use of this material is permitted. Permission from IEEE must be obtained for all other uses, in any current or future media, including reprinting/republishing this material for advertising or promotional purposes, creating new collective works, for resale or redistribution to servers or lists, or reuse of any copyrighted component of this work in other work}
\newcommand\copyrightnotice{%
\begin{tikzpicture}[remember picture,overlay]
\node[anchor=south,yshift=10pt] at (current page.south) {\fbox{\parbox{\dimexpr\textwidth-\fboxsep-\fboxrule\relax}{\copyrighttext}}};
\end{tikzpicture}%
}

\title{\LARGE \bf
Is attention to bounding boxes all you need for pedestrian action prediction?
}

\author{Lina Achaji$^{1,2}$, Julien Moreau$^{1}$, Thibault Fouqueray$^{1}$, Francois Aioun$^{1}$, and Francois Charpillet$^{2}$
\thanks{$^{1}$Stellantis Group, Technical center of Velizy 78140, France
{\tt\small lina.achaji@stellantis.com}
{\tt\small julien.moreau@stellantis.com}
{\tt\small thibault.fouqueray@stellantis.com}
{\tt\small francois.aioun@stellantis.com}}
\thanks{$^{2}$Université de Lorraine, CNRS, Inria, LORIA, Nancy 54000, France
        {\tt\small francois.charpillet@inria.fr}}%
}

\begin{document}

\maketitle
\copyrightnotice
\thispagestyle{empty}
\pagestyle{empty}

\begin{abstract}

The human driver is no longer the only one concerned with the complexity of the driving scenarios. Autonomous vehicles (AV) are similarly becoming involved in the process. Nowadays, the development of AV in urban places raises essential safety concerns for vulnerable road users (VRUs) such as pedestrians. Therefore, to make the roads safer, it is critical to classify and predict the pedestrians' future behavior. In this paper, we present a framework based on multiple variations of the Transformer models able to infer predict the pedestrian street-crossing decision-making based on the dynamics of its initiated trajectory. We showed that using solely bounding boxes as input features can outperform the previous state-of-the-art results by reaching a prediction accuracy of 91\% and an F1-score of 0.83 on the PIE dataset. In addition, we introduced a large-size simulated dataset (CP2A) using CARLA for action prediction. Our model has similarly reached high accuracy (91\%) and F1-score (0.91) on this dataset. Interestingly, we showed that pre-training our Transformer model on the CP2A dataset and then fine-tuning it on the PIE dataset is beneficial for the action prediction task. Finally, our model's results are successfully supported by the ``human attention to bounding boxes'' experiment which we created to test humans ability for pedestrian action prediction without the need for environmental context. \textit{The code for the dataset and the models is available at: \url{https://github.com/linaashaji/Action_Anticipation}} 

\end{abstract}

\section{INTRODUCTION}

During the course of our lives, we observe the movement of objects around us and simultaneously simulate their future trajectory. Knowing whether this object is moving fast or slow affects these predicted trajectories accordingly \cite{barsalou2009simulation}. The transition of our mobility system into the era of autonomous driving is often seen as adding a layer of artificial intelligence to conventional vehicle platforms \cite{BEHERE2016136}. It should be capable of not only perceiving the world, but also predicting and analyzing its future states. The predictive processing can be applied everywhere, especially when interacting with vulnerable road users (VRU) such as pedestrians. When dealing with pedestrians, we can formulate the prediction as a higher-level semantic prediction such as the early anticipation of the future action, for example, walking, running, performing hand gestures, or most importantly crossing or not crossing the street in front of the AV. Recently, trajectory and action prediction solutions have been proposed based on sequential reasoning that mainly use algorithms built on recurrent neural networks (i.e., RNN, LSTM) \cite{rasouli2019pie, pop2019multi, kotseruba2021benchmark, poibrenski2020m2p3}. However, it has recently became clear that LSTM lacks many capabilities to model sequential data. For instance, LSTM suffers from long-term prediction, often due to the vanishing gradient problem \cite{sutskever2014sequence}. That leads to its inability to model the correlation between non-neighboring inputs in the sequence. Furthermore, during training, LSTM is not able to assign different weights to different tokens based on their relative importance to the output. This will force it to give equal attention to all inputs, even if the capacity of the model does not allow it.  
Hence, the attention coupled with LSTM \cite{bahdanau2014neural} has enhanced the previously suggested solution by proposing a mathematical framework that can weight each of the input tokens differently depending on their importance to the input sequence itself (self-attention) \cite{lin2017structured} or the output sequence (cross-attention). Nevertheless, attention mechanisms coupled with LSTM have limited the potential of the attention framework itself. We can see that using only attention, i.e., Transformer architecture \cite{vaswani2017attention} can lead to better results. Transformers have first revolutionized the natural language processing problems by outperforming all the previously proposed solutions \cite{lagler2013gpt2, devlin2018bert}. However, they have recently proved to be equally efficient for non-NLP problems \cite{neimark2021video, giuliari2021transformer, aksan2020attention}. 

In this paper, we leverage Transformer Networks for the sake of predicting the future pedestrian street crossing decision-making. In fact, This network, fed solely with bounding boxes as input features, will predict the crossing-probability of a pedestrian in front of the ego-vehicle using an observation sequence of 0.5 seconds and with a prediction horizon varying between one and two seconds in the future. To achieve this goal, We propose multiple variants of the Transformer architecture: the Encoder-only, the Encoder coupled with pooling-layers, and the Encoder-Decoder.
Moreover, we generated The CP2A dataset using the CARLA simulator and used it along with the PIE dataset \cite{rasouli2019pie} to evaluate our models' performance. Additionally, we will show that the pre-trained weights on the CP2A dataset are capable of boosting the performance of our fine-tuned models on the PIE dataset.

\section{Related Work}
In this section, we review the recent advances in pedestrian action and trajectory  prediction. Then, we discuss the work done using Transformers in various field of applications and the effect of Transfer learning on Transformer models.

\noindent \textbf{Pedestrian Action Anticipation} is a highly important problem for autonomous cars. Where the objective is to anticipate in advance the possibility of a pedestrian to cross the road in front of the ego-vehicle. This problem was addressed using multiple approaches. For instance, \cite{pop2019multi} used a network of convolutional neural networks and LSTM to detect the pedestrians and predict their actions up to 1.3 seconds ahead in the future and their estimated time-to-cross. The model in \cite{rasouli2020pedestrian} used a stacked GRU network composed of five GRUs each of which processes a concatenation of different features (i.e., the pedestrian appearance, the surrounding context, the skeleton poses, the coordinates of the bounding boxes, and ego-vehicle speed). The authors in \cite{gesnouin2020predicting} converted the human pose skeleton sequences into 2D image-like spatio-temporal representations and then applied CNN-based models. Lately, a benchmark paper \cite{kotseruba2021benchmark} was published to evaluate action prediction on the PIE \cite{rasouli2019pie} and JAAD \cite{rasouli2017they} datasets. 

\noindent \textbf{Pedestrian Trajectory Prediction} is a closely related task to action prediction. In contrast, the output sequence is a set of predicted positions in the future. In recent works, we see that intention and action prediction can be critical for trajectory prediction \cite{rasouli2019pie}. The M2P3 model \cite{poibrenski2020m2p3} used an Encoder-Decoder RNN along with conditional variational auto-encoder (CVAE) to predict multi-modal trajectories from an ego-centric vehicle view. GRIP++ \cite{li2019grip} builds a dynamic GNN (Graph Neural Network) to model the interaction between agents in the scene. The Trajectron \cite{ivanovic2019trajectron} combines elements from CVAE, LSTM, and dynamic Spatio-temporal graphical structures to produce multimodal trajectories. Recently, a Transformer model was proposed by \cite{giuliari2021transformer} to predict the future trajectories of the pedestrians conditioning on the previous displacement of each pedestrian in the scene. The transformer used has the same architecture as the Vanilla Transformer proposed in \cite{vaswani2017attention}.

\noindent \textbf{Transformers Networks} are self-attention-based models proposed by \cite{vaswani2017attention} for machine translation and have since became the state-of-the-art method in many NLP tasks. Here, we focus on the development of Transformers for modeling actions, motion dynamics, and visual context. For instance, \cite{girdhar2019video} introduced an Action Transformer model for human action localization and recognition in video clips. The authors in \cite{aksan2020attention} proposed a Spatio-temporal transformer for 3D human motion modeling by learning the evolution of skeleton joints embeddings through space and time. Recently, \cite{bertasius2021space} and \cite{arnab2021vivit} applied a Spatio-temporal transformer for video action recognition. Similar to the NLP-oriented Transformers, where Large Transformer-based models are often pre-trained on large datasets and then fine-tuned for a particular task, \cite{bertasius2021space, arnab2021vivit} used a pre-trained Image Transformer \cite{dosovitskiy2020image} weights, bootstrapped the weights, and then fine-tuned the model on a very large dataset \cite{carreira2017quo} for action recognition. This training on the large dataset has remarkably increased the performance of Transformers for visual action recognition.

\section{Method}

The prediction and analysis block in autonomous car architectures receives its inputs from the perception block. We assume that the sets of observations of every pedestrian agent are arranged in sequences. Every sequence describes their states between a time-steps $t=m$, and $t=M$: $O_{m:M} = \{O_m, \dots, O_M\}$.
Conditioning on the whole history of observations, the objective is to calculate the following probability distribution:
$$ p(\:P_{a:A} \:| \:O_{m:M}\:)$$
where $P_{a:A} = \{P_{a} , \dots, P_{A} \} $, such that $P_t$ is the predicted state of a particular pedestrian at time $t$.
For the early action anticipation, $a$ is equal to $A$. In fact, if the pedestrian crosses the street, $A$ represents the starting time of the crossing event. Otherwise, it represents the time of the pedestrian's last observable frame. The time between the last observation frame $M$ in the observation sequence and the critical time $A$ is called the Time-To-Event (TTE).

\begin{figure*}
  \includegraphics[scale=0.28]{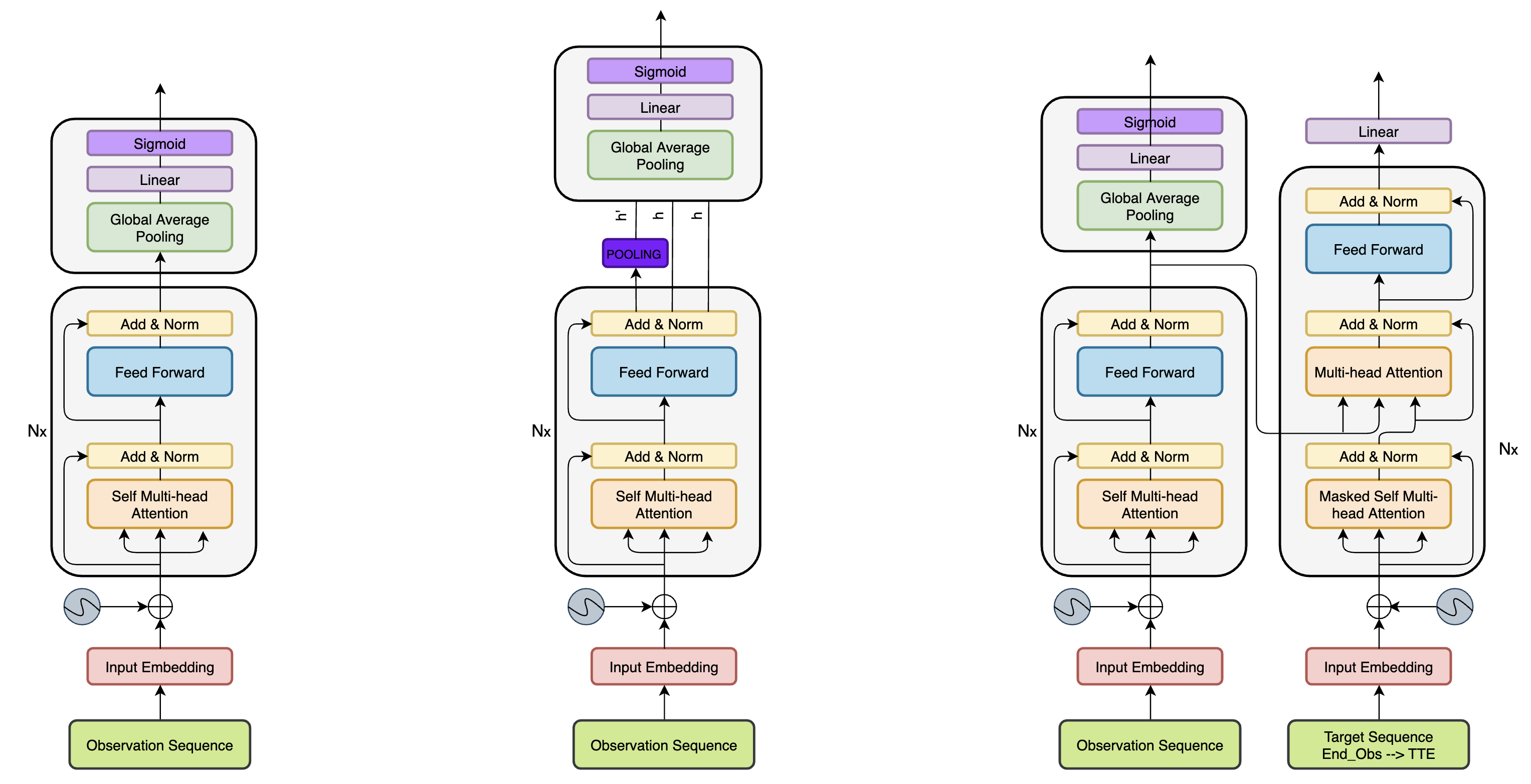}
\centering
\caption{(Left) Encoder-Only; (Middle) Encoder-Pooling; (Right) Encoder-Decoder. Varied from \cite{vaswani2017attention}.}
\label{fig:archi}
\end{figure*}

The input to our model is composed of the bounding box coordinates of each pedestrian in the scene. The bounding box is represented by the x,y coordinates of the upper-left corner and the such coordinates of the lower-right corner of the rectangle around the pedestrian, without the use of images or visual context. Certainly, these bounding boxes were first extracted from images, either manually (i.e., in the case of annotated datasets) or using state-of-the-art object detectors. 

In the following section, we will present our Transformer Network based model.
\subsection{Input Embedding}

Before feeding the input $X_T \in \R^{T \times 4}$ to our encoder, we first project the 4-dimensional bounding boxes into a D-dimensional space via a linear layer. 
$$E_T = X_T W_e + b_e$$
such that, $E_T \in \R^{T \times D}$, $W_e \in \R^{4 \times D}$, and $b_e \in \R^{T \times D}$.  $W_e$ and $b_e$ correspond respectively to the learnable weight and the bias that project the bounding boxes vector into the embedding space.

Transformers deal with unordered sets instead of sequences. Therefore, we follow \cite{vaswani2017attention} by injecting fixed sinusoidal positional encoding into the input embeddings.

\subsection{Attention Layer}
An attention layer updates each component of a sequence by aggregating global information from the complete input sequence. This is done by defining three learnable weight matrices to transform Queries ($W_Q \in \R^{D \times d_q}$), Keys ($W_K \in \R^{D \times d_k}$), and Values ($W_Q \in \R^{D \times d_v}$). The output  of the attention layer is then given by:

$$ Attention(Q, K, V, M) = softmax(Q K^T / \sqrt(d_k) + M )V $$

Where the mask M prevents information leaking from future steps. Following \cite{vaswani2017attention}, we project the D-dimensional embedding space using the multi-head attention framework into $H$ subspaces calculated by the multiple attention heads $i \in \{1,\dots ,H \}$:

$$ MHA(Q^i, K^i, V^i) = Concat( head_1, \dots, head_h) W^O$$ 

Where $Q^i$ , $K^i$, and $V^i \in \R^{T \times F}$. We set $F = D/H$. The $head_i = Attention(QW_i^Q, KW_i^K, VW_i^V)$, and $W^O \in \R^{D \times D}$.

\subsection{Transformer Architecture}

We proposed to apply multiple variations of the Transformer model:

\subsubsection{Encoder-Only Architecture (TEO)}
The encoder-only model goes into a stack of $N$ encoder layers. Each encoder layer consists of a multi-head self-attention layer and point-wise feed-forward neural networks (P-FFN). We define $h_n \in \R^{T \times D}$ as the output of the $n-$th encoder layer:

$$h_n = LayerNorm(h_{n-1} + MHA(Q = h_{n-1}, KV = h_{n-1}))$$
$$h_n = LayerNorm( h_n + P-FFN(h_n))$$

The output sequence length of the encoder layers remains the same as the input $\in \R^{T \times D}$. However, to predict the pedestrian action decision, we compress the output sequence in the temporal domain by applying a global average pooling layer at the top of the encoder stacks. Then we apply another embedding layer followed by a sigmoid function to get the crossing action probability.

\subsubsection{Encoder-Pooling Architecture (TEP)}
Instead of applying the global average pooling at the top of the encoder stack of layers directly, we proposed pooling intermediate layers between the encoder blocks (Fig. \ref{fig:archi} middle). We see this solution as a compact way to reduce the observation sequence length gradually from the input to the output as the layer gets deeper. This solution will prevent the network from directly transforming the last layer embedding size from $\R^{T \times D}$ to $\R^{1 \times D}$ to predict the one-dimensional action.  Instead, the final sequence length will be $T'$, a reduced transformation of the initial sequence size $T$. In fact, pooling layers have been used in CNN architectures to reduce the dimensions of the feature maps. Similarly, \cite{dai2020funnel} proposed to adapt the pooling layers to achieve representation compression and computation reduction in Transformer networks. At the output of each encoder layer, we will apply a strided mean pooling layer on the query vector to reduce its sequence length:

$$h'_{n-1} = pooling(h_{n-1}),$$
where $h' \in \R^{T' \times D}$ for $T' < T$,  and the unpooled sequence h serves the role of the key and value vectors:
$$h_{n} = LayerNorm(h_{n-1} + MHA(Q = h'_{n-1}, KV = h_{n-1}))$$

\subsubsection{Encoder-Decoder Architecture (TED)}
The final architecture is the encoder-decoder. The encoder block is identical to the encoder-only architecture with the classification head on the top of the encoder stacks. For the decoder architecture, we input the target sequence $T_{a:A} = Y_C \in \R^{C \times 4}$, where $a$ is equal to the next-observation time frame $M+1$, and $A$ is equal to the critical time TTE. Each decoder layer consists of a masked multi-head self-attention layer, a masked multi-head cross-attention layer, and point-wise feed-forward neural networks (P-FFN). The cross-attention layer takes the query from the previous self-attention layer along with the keys and the values from the encoder memory: 

\begin{multline*}
h_{dec_{n}} = LayerNorm(h_{dec_{n-1}} + \\
MHA(Q = h_{dec_{n-1}}, KV = h_{enc_{N}}))
\end{multline*}

The output of the decoder block is the target input shifted by one representing the future trajectory of the pedestrian between timesteps $t=a+1$ and $t=A+1$. This architecture will jointly learn to classify the pedestrian's future action as well as its future trajectory. In the following sections, we will see that the action prediction performance is increased by training the Transformer to jointly learn the pedestrian crossing action and trajectory.
\subsection{Training and Inference}
We train the TEO and the TEP models by optimizing the classification binary cross entropy (BCE) between the predicted and target class. For the TED model, we use a combined weighted loss of the BCE and the $l_2$ distance between the predicted and target sequence:

$$ L_{TED} = \lambda_{cls} \: BCE + \lambda_{reg} \: l_2(Y_{C+1}, \hat{Y_{C+1}})$$ 
Where $\lambda_{cls}$ and $\lambda_{reg}$ are the classification and regression hyperparameters to be tuned.

\section{Experiments}
In this section, we evaluate our proposed models on two datasets following the evaluation protocols in \cite{kotseruba2021benchmark}. We compare our model results with the baselines based on different feature choices. We fix the observation length for all models at 16 frames (i.e., 0.5 seconds) and the Time-To-Event (TTE) between 30 and 60 frames (i.e., 1 and 2 seconds). We examine different settings of model structure (Section \ref{ablation}) and explore the effect of changing the temporal prediction horizon on the results. Also, we investigate the effect of transfer learning on our models.

\begin{table}[b]
\caption{Comparison of the CP2A dataset with other pedestrian action prediction datasets.}
\centering
\begin{tabular}{ c c c c c}
 \toprule[.1em]
 Dataset & Running Time & \# Bounding Boxes & S/R\\
\midrule[.1em]
 PIE \cite{rasouli2019pie} & 6 hours & 740 K & Real\\
\midrule[.1em]
 JAAD \cite{kotseruba2016joint} & 1.5 hours & 391 K & Real\\
\midrule[.1em]
 STIP \cite{liu2020spatiotemporal} & 15.4 hours & 3.5 M & Real\\
 \midrule[.1em]
 CP2A (ours) & 5.5 days & 14 M & Simulated\\
 \bottomrule[.1em]
\end{tabular}
\label{dataset}
\end{table}

\subsection{Datasets}
\subsubsection{Pedestrian Intention Estimation (PIE) dataset}
The PIE dataset \cite{rasouli2019pie} provides 6 hours of continuous footage recorded at 30 frames per second (FPS) in clear weather conditions. The dataset contains different pedestrian behaviors near the crossing event time. Also, it includes urban environments with various traffic flow intensities. For each pedestrian who can interact with the ego-vehicle driver, it provides the coordinates of the bounding boxes, the critical time where each pedestrian will cross, as well as their actions of crossing or not crossing the street.

\begin{figure}[b]
\includegraphics[scale=0.32]{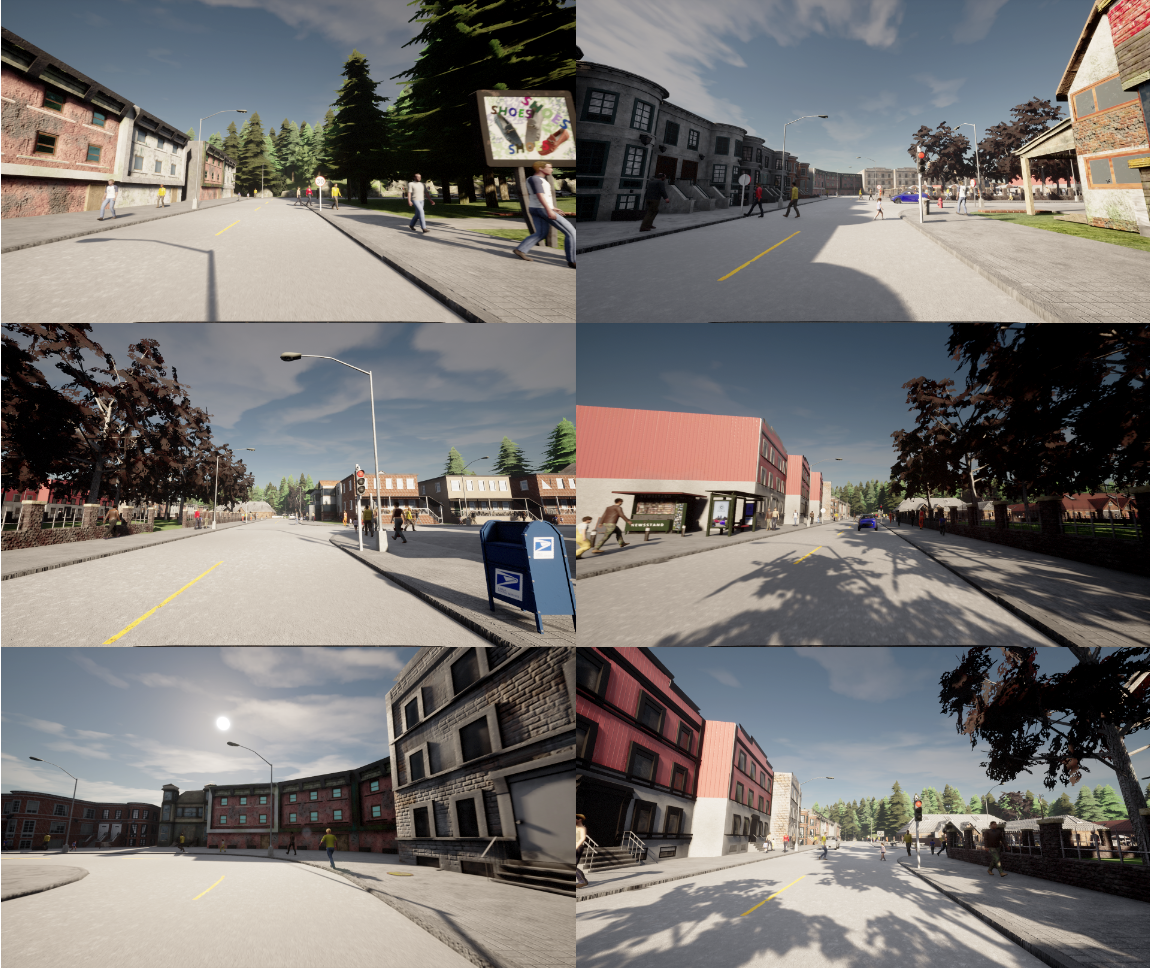}
\centering
\caption{Samples from the simulated CP2A dataset.}
\label{fig:cp2a}
\end{figure}

\begin{table*}[t]
\caption{Accuracy, Area Under Curve (AUC), and F1-score comparison with baseline models. TEO: Transformer Encoder-only, TEP: Transformer Encoder-Pooling, TED: Transformer Encoder-Decoder, FTEO: Fine-tuned Transformer Encoder-only, BB: Bounding Boxes, P: Pose-Skeletons, S:  Ego-vehicle Speed, I: RGB Images. \textbf{*:} Trained on the CP2A dataset.} 
\centering
\begin{tabular}{ c c c c c c}
 \toprule[.1em]
 Model Name & Architecture & Features & Accuracy & AUC & F1-score\\
\midrule[.1em]
 Static   & \multirow{3}{*}{CNN} & \multirow{2}{*}{I}    & 0.71 &   0.60 &   0.41\\
 \multirow{2}{*}{I3D} & & & 0.80   & 0.73 & 0.62\\

  &  & Optical flow & 0.81   & 0.83 & 0.72\\
\midrule[.1em]
 MultiRNN & GRU & \multirow{2}{*}{BB, P, S, I} & 0.83 &  0.8 &  0.71\\
 PCPA & LSTM + Attention & & 0.87   & 0.86 & 0.77\\
\midrule[.1em]
CAPformer & Transformer & BB, S, I & - & 0.853 & 0.779\\
\midrule[.1em]

 PCPA & LSTM + Attention & \multirow{4}{*}{BB} & 0.48   & 0.42 & 0.57\\
 TEO (Ours) & \multirow{3}{*}{Transformer} &  & 0.88   & 0.85 & 0.77\\
 TEP (Ours) &  &  & 0.88   & 0.87 & 0.77\\
 TED (Ours) &  &  & \textbf{0.91}   & \textbf{0.91} & 0.83\\
 \midrule[.1em]
  \midrule[.1em]
 FTEO (Ours) & Fine-Tuned Transformer & BB & 0.89   & 0.89 & \textbf{0.88}\\
  \midrule[.1em]
 CP2A\textbf{*} (Ours) & Transformer & BB & 0.9   & 0.9 & 0.9\\
 \bottomrule[.1em]
\end{tabular}
\label{results}
\end{table*}

\subsubsection{CARLA Pedestrian Action Anticipation (CP2A) dataset}
In this paper, we present a new simulated dataset for pedestrian action anticipation collected using the CARLA simulator \cite{Dosovitskiy17}. Indeed, Transformer networks have shown an advantage over other types of algorithms when a large amount of training data is available \cite{dosovitskiy2020image}. Moreover, pre-training Transformer networks on large datasets followed by a fine-tuning procedure has proven to be very effective \cite{bertasius2021space}. To this end, we have simulated a very large dataset (see Table \ref{dataset}) that can be automatically and freely labeled using the simulator itself. Using such dataset, we can first evaluate our model and then improve the overall performance using transfer learning on the real collected scenarios such as the PIE dataset. To generate this dataset, we place a camera sensor on our ego-vehicle and set the parameters to those of the camera used to record the PIE dataset (i.e., 1920x1080, 110° FOV). Then, we compute bounding boxes for each pedestrian interacting with the ego vehicle as seen through the camera's field of view. We generated the data in two urban environments available in the CARLA simulator: Town02 and Town03.
The total number of simulated pedestrians is nearly 55k, equivalent to 14M bounding boxes samples. The critical point for each pedestrian is their first point of crossing the street (in case they will eventually cross) or the last bounding box coordinates of their path in the opposite case. The crossing behavior represents 25\% of the total pedestrians. We balanced the training split of the dataset to obtain labeled sequences crossing/non-crossing in equal parts. We used sequence-flipping to augment the minority class (i.e., crossing behavior in our case) and then undersampled the rest of the dataset. The result is a total of nearly 50k pedestrian sequences.
Next, following the benchmark in \cite{kotseruba2021benchmark}, the pedestrian trajectory sequences were transformed into observation sequences of equal length (i.e., 0.5 seconds) with a 60\% overlap for the training splits. The TTE length is between 30 and 60 frames. It resulted in a total of nearly 220k observation sequences. Samples from the CP2A dataset are shown in (Fig. \ref{fig:cp2a}). We used an Nvidia 1080 Ti GPU for data simulation with a generation rate of about 600 pedestrian sequences per hour, equivalent to the size of the PIE dataset per 2 hours.  

\subsection{Baseline Models}
Recently, the authors in \cite{kotseruba2021benchmark} have published a benchmark on the PIE dataset for evaluating pedestrian action prediction. They unified multiple baseline models under the same evaluation procedures.
\subsubsection{Static}It is a model that predicts the action based on the VGG16 backend \cite{simonyan2014very} using only the last time step frame in the observation sequence.
\subsubsection{Multi-stream RNN (MultiRNN) \cite{bhattacharyya2018long}} it is composed of separate GRU streams independently processing the following feature types: Bounding Boxes coordinates, the Pose Skeletons, the Ego-vehicle speed, and the local box frame cropped around the Bounding Box that is processed by a VGG16 backbone. 
\subsubsection{Inflated 3D (I3D) network \cite{carreira2017quo}} It predicts the action by taking a stack of RGB images corresponding to a video into a multi-stream 3D CNN network followed by a fully connected layer.
\subsubsection{PCPA \cite{kotseruba2021benchmark}} it is composed of multiple RNN branches to encode non-visual features (e.g., Bounding Boxes coordinates, the Pose Skeletons, and the Ego-vehicle speed) along with a C3D network to encode each pedestrian's local context. The outputs of the branches are then fed into modality attention layers.
\subsubsection{CAPformer \cite{capformer}} it is a transformer-based architecture composed of various branches fusing raw images and kinematic data. The video branch uses the Timesformer \cite{bertasius2021space} backbone. 

\subsection{Results on PIE dataset}

Table \ref{results} shows the comparison of the results of our proposed models with the state-of-the-art baselines on the PIE dataset. 
As seen, our Transformer models TEO, TEP, and TED based solely on Bounding Boxes features outperform the SOTA baselines that use Bounding Boxes alongside Pose-skeletons, Ego-vehicle speed, and RGB images as features. Additionally, we trained the PCPA model with only Bounding Boxes as input, similarly to our model. In this case, the PCPA model results dropped dramatically and prevented the model from learning any significant pattern about the action prediction task. This finding can show the superiority of Transformers compared to LSTMs.  
Recently, the authors of CAPformer \cite{capformer} have used spatio-temporal and temporal transformers by merging the features of bounding boxes, ego vehicle speed, and raw images. Our models also outperform their results. The drop in their proposed model performance is due to the non-maturity of Transformer models when trained on images with complex or chaotic scenarios. Thus, further research should be conducted here to analyze the best techniques for capturing the necessary features of these images without falling into the curse of dimensionality, especially when dealing with short-duration videos (0.5 seconds) and relatively low-resolution images. 
Lastly, as shown in Table \ref{results}, the TED's model results exceeded the TEO and TEP ones in terms of ACC (92 \%), F1-score (0.86), and AUC (0.9). Based on these results, we can show that training the model jointly on trajectory prediction and action prediction is better then training solely on the action prediction task.

\subsection{Ablations}\label{ablation}

\begin{figure}[b]
\includegraphics[scale=0.23]{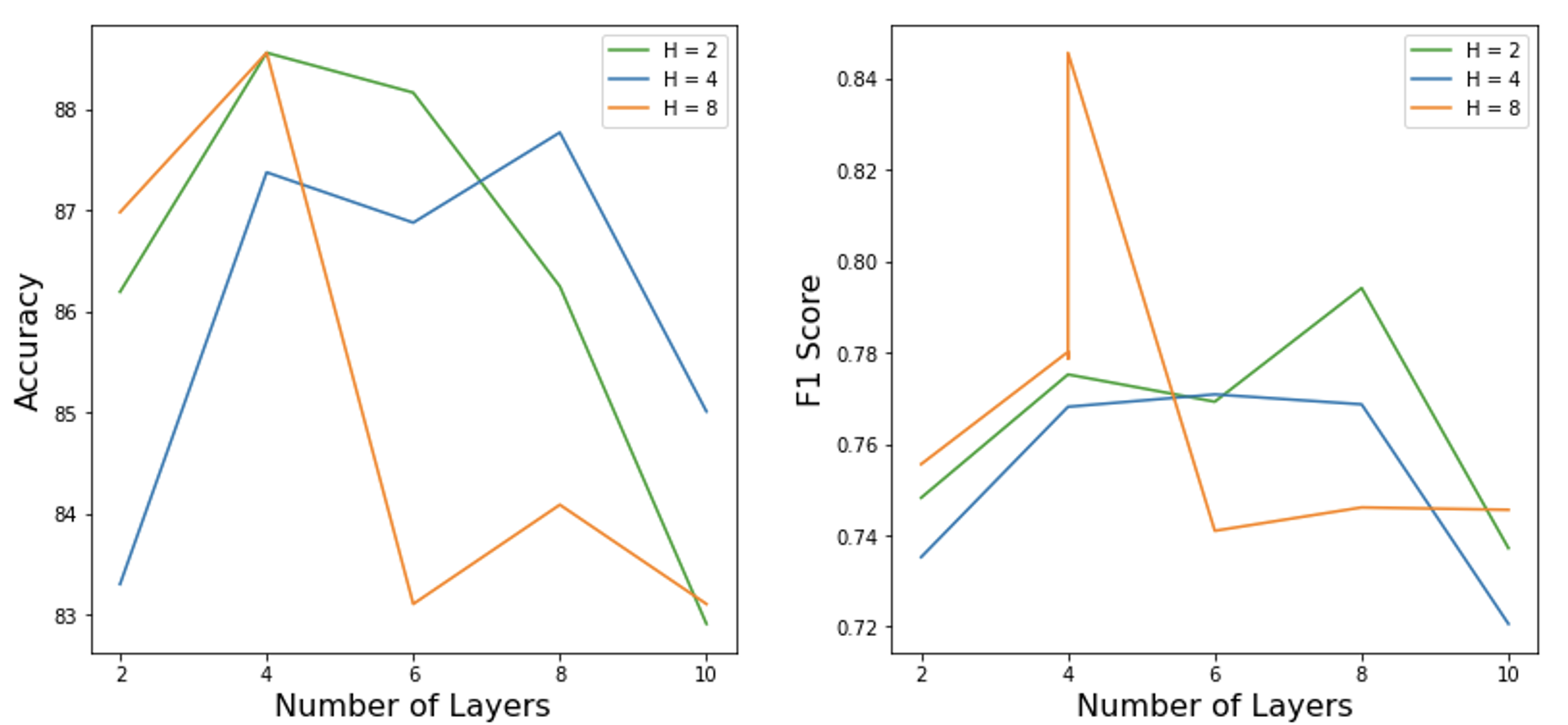}
\centering
\caption{Performance of the TEO model  when trained with different number of attention heads  and attention layers.}
\label{fig:metrics}
\end{figure}

\begin{figure}[t]
\includegraphics[scale=0.23]{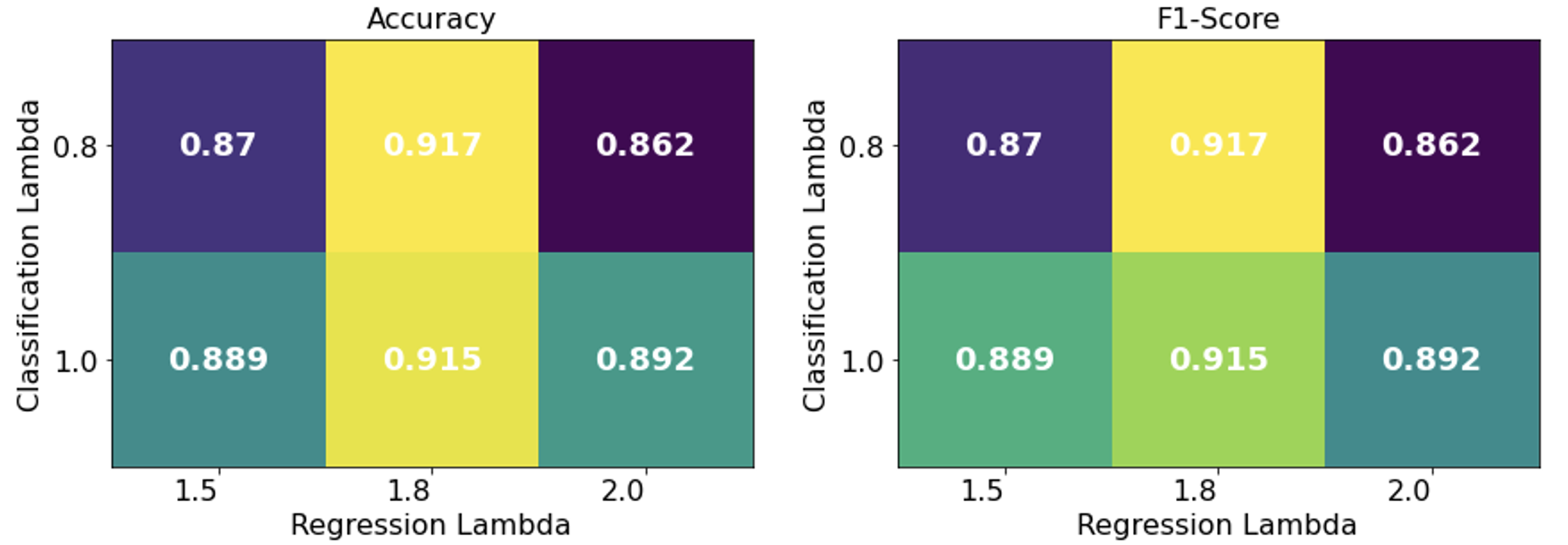}
\centering
\caption{Performance of the TED model when trained with different regression and classification hyper-parameters.}
\label{fig:metrics_ted}
\end{figure}

Figure \ref{fig:metrics} shows the performance of our TEO model when trained with different numbers of layers and heads. We obtained the best performance for the 4-layer and 8-head configurations (shown in Table \ref{results}) and 8-layer, 2-head configurations. For the TEP and TED, the models reach their best performance when considering the 8-layer and 8-head settings. Regarding the TED model, we experiment with different hyperparameters for the regression and classification values in the TED loss function settings. We obtain the best model (Fig. \ref{fig:metrics_ted}) using the (1.8, 0.8) combination for the regression and classification parameters respectively. In our experiments, we use an embedding dimension $D$ of 128, a P-FFN hidden dimension $Dff$ of 256, and batch size of 32. We use the Adam \cite{kingma2014adam} optimizer with a learning rate of $10^{-4}$. Confirming the findings in \cite{liu2020understanding}, we noticed that using stochastic optimizers for our Transformer models is not stable, and the results changed remarkably from one experiment to another using the same hyper-parameters. 

\begin{figure}[t]
\includegraphics[scale=0.26]{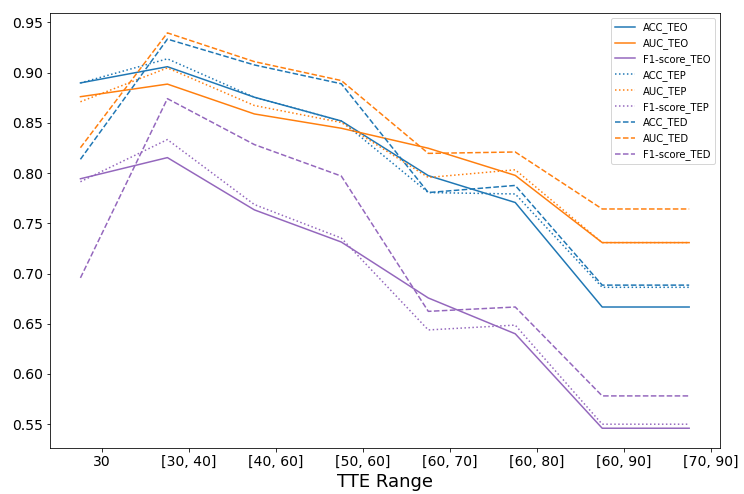}
\centering
\caption{Changing the prediction time horizon range for the models trained on the [30, 60] frames TTE interval.}
\label{fig:tte}
\end{figure}

We study the prediction performance variance when modifying the temporal prediction horizon (equivalent to TTE time). We set the length of the observation sequence to 16 frames (i.e., 0.5 s) and varied the TTE time from 0.5 seconds to 3 seconds in the future (90 frames). These results were reported on the test set by taking the same models that were trained on the original 30-60 frame TTE interval. The graphs (Fig. \ref{fig:tte}) show that our models reach their upper limit when predicting the upcoming 1 to 1.3 s interval with an accuracy of 93\% for the TED model. Furthermore, the model still performs reasonably well even when all prediction horizons are between 2 and 2.3 seconds, with nearly 80\% accuracy for all three models. We performed all our experiments on an Nvidia 1080 Ti GPU. We reported the inference time (Table \ref{inftime}) of the three proposed models using their best hyperparameters scenarios. We should note here that we can use only the encoder block of the TED model in the inference phase, without the need to predict the future pedestrian trajectories. Hereby, reducing the model's inference time.

\begin{table}[b]
\caption{Inference Time on Nvidia 1080 Ti GPU.}
\centering
\begin{tabular}{ c c c c}
 \toprule[.1em]
 Model Name & Number of Layers & Inference Time (ms)\\
\midrule[.1em]
 TEO & 4  & 1.63\\
\midrule[.1em]
 TEP & 8 & 2.85\\
\midrule[.1em]
 TED & 8 & 2.76\\
 \bottomrule[.1em]
\end{tabular}
\label{inftime}
\end{table}

\subsection{Results on CP2A}

In general, the action prediction task is related to understanding the environmental visual context. Our model fed with solely bounding boxes as input showed controversial results by outperforming models trained with environmental content as context features. Hence, it is relevant to question if the PIE dataset is biased. Although previous models were not able to detect these patterns, Transformers were able to find this bias. To test this hypothesis, we trained our models on the CP2A dataset, which is 19x larger than the PIE dataset and is highly unlikely to have the same bias that can be found in the PIE dataset. The raw CP2A in Table \ref{results} shows the performance of the 8-layer, 8-head TEO model on the simulated dataset. It achieved 90\% accuracy and 0.9 for AUC and F1-score, demonstrating the effectiveness of using transformers with bounding boxes, regardless of the choice of dataset. 

\subsubsection{Effect of Transfer Learning}
Interestingly, we obtained outstanding results when using the pre-trained weights from the CP2A model and fine-tuning them on the TEO model using the PIE dataset (FTEO model in Table \ref{results}). In particular, we can see the performance gain (+0.11) in terms of F1-score, where we reached 0.88, which outperforms the TED model trained from scratch on PIE, and slightly outperforms the baseline TEO model in terms of accuracy and AUC. In brief, the transfer of knowledge regarding pedestrian action prediction is effective. The results obtained here are consistent with the advantage observed in computer vision \cite{bertasius2021space} when applying transfer learning from larger datasets.

\section{Human Attention To Bounding Boxes Experiment}

Our Transformer based model, fed with bounding boxes as input, shows controversial results since it outperformed other models that were trained with environmental context as input features. To ensure that our results are bias-free, we conducted a survey \footnote{\url{https://www.psytoolkit.org/c/3.3.2/survey?s=LT99F}. We created the survey using \cite{stoet2010psytoolkit,stoet2017psytoolkit}. The survey was completely online and the link was distributed among the Inria Nancy research lab (Larsen team) students and employees. The participants were anonymous volunteers who responded to the survey.} to test the humans' ability to predict future pedestrians' street-crossing decision-making by considering only the trajectory of the bounding boxes. We should note here that the purpose of our study is different from the paper in \cite{Schmidt} that examined if humans are better at predicting actions using environmental context or only trajectory dynamics. However, we target a different purpose by examining whether humans are capable of predicting future pedestrian actions without environmental context, alike transformers.

\begin{figure}[t]
\includegraphics[scale=0.1]{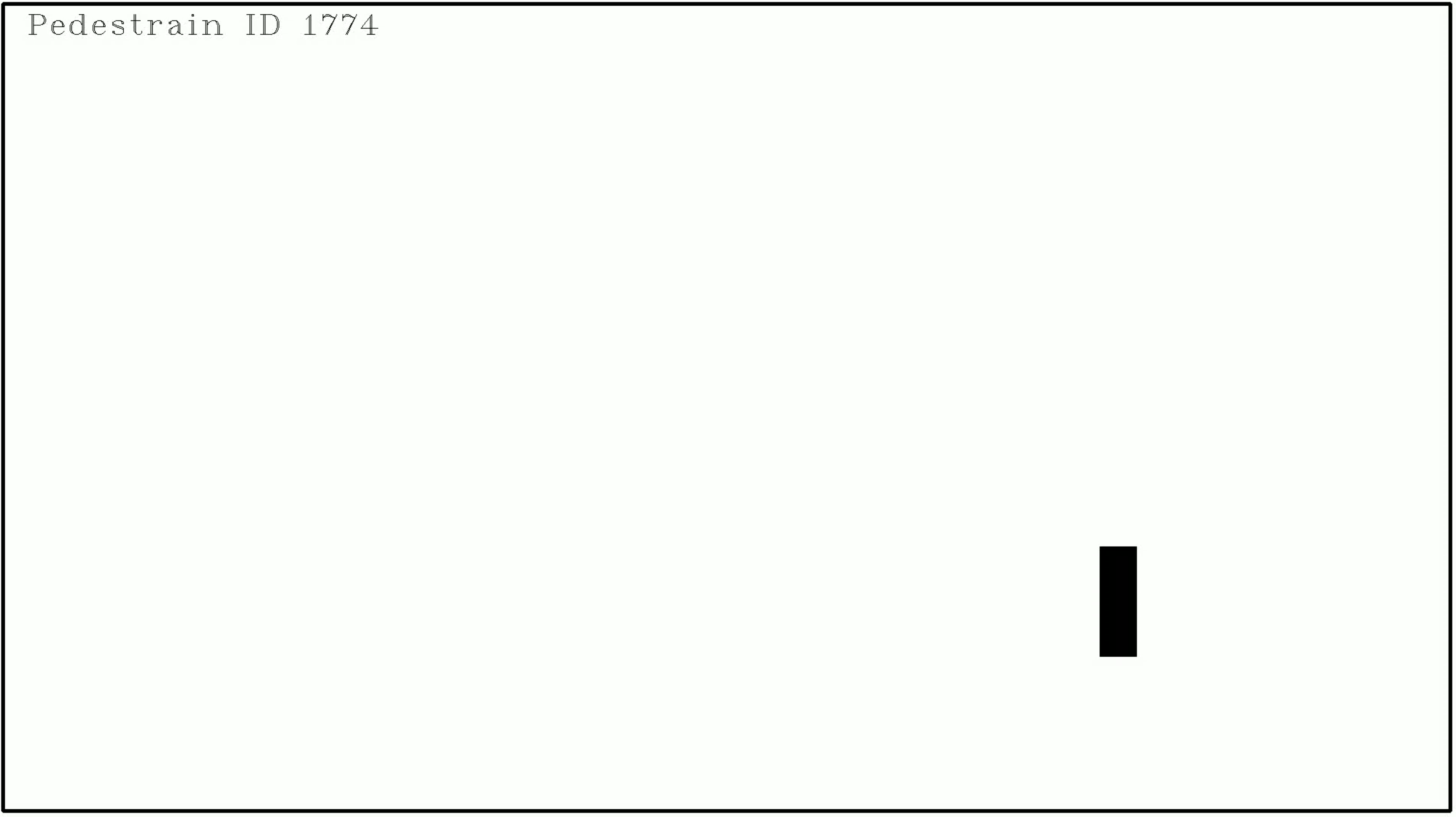}
\centering
\caption{A sample from the testing videos of the human attention to bounding boxes experiment.}
\label{fig:human}
\end{figure}

\subsection{Survey Protocol}
During the survey, we will show the participants a series of white videos with a black rectangle representing the shape of the bounding box and the original motion trajectory of the pedestrian (Fig. \ref{fig:human}). The observation sequence length is 0.5 seconds, as used in the Transformer models. The participants should predict the crossing decision that should occur after 1.5 seconds in the future. The experiment is split into three main phases:

\begin{itemize}
    \item In the first phase, the participants will see a regular video from the PIE dataset of pedestrians in front of a vehicle. The camera location will always be the same. For each pedestrian, we are showing the trajectory of its bounding box for 0.5 seconds. If the pedestrian crosses the street in the future (after 1.5 seconds), the bounding box will be colored green. Otherwise, it will be red.  
    \item In the second phase (Training phase), we are concerned by training the participants to get familiar with the experience. Instead of the original image, the participants will see multiple examples of pedestrians' bounding boxes filled in black, where the rest of the image (i.e., environment, other pedestrians, vehicles) will be masked in white. They should try to predict the pedestrian crossing decision in front of the (hidden) ego-vehicle. During this phase, we will show the participants the correct answers after each observation sequence. 
    \item In the next phase (Testing phase), we will test the participants with different examples. They will be asked to answer the binary prediction question (cross/not cross). 
\end{itemize}
At the end of the experiment, the participants will be asked to fill in the chosen features for their action prediction answers.

The training phase video series is composed of 26 videos, each consisting of 16 frames (0.5 seconds) of pedestrian observation sequence. The 26 samples are split equally between positive (crossing) and negative (not crossing) behavior. 
The testing phase is composed of 44 videos of 16 frames each. Similarly, the crossing/not crossing distribution is 50\% each. It is worth noting here that the participants do not know the distribution beforehand. The final score of each participant will be calculated based on the testing videos.

\begin{figure}[t]
\includegraphics[scale=0.4]{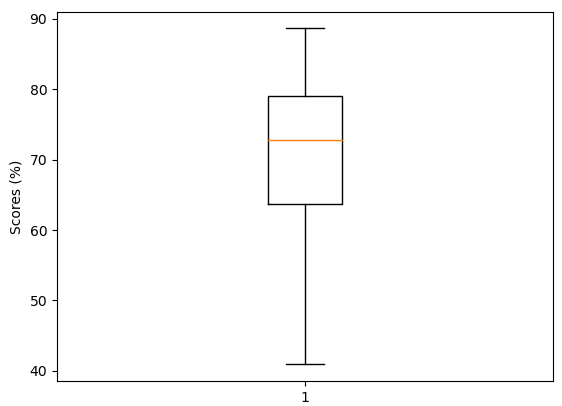}
\centering
\caption{Box Plot showing the score distribution with a mean of 70\% and a maximum of 88\%.}
\label{fig:boxplot}
\end{figure}

\subsection{Results and Analysis}

A total of 26 participants joined the experiment. The results obtained interestingly confirm our hypothesis regarding the Transformer model. Using only the bounding box features and hiding everything else, we obtained an accuracy of 70\% on average among the participants. Furthermore, 24\% of the participants scored above 80\%, with a maximum score of 88\%, which is close to the Transformer score (Fig. \ref{fig:boxplot}). These results can reassure us that with proper training, humans and Transformers can capture models to predict pedestrian actions without the need for the environment. This result does not directly imply that humans and Transformers use the same patterns for prediction, but it can, at least, show that such models exist and can be used effectively. The main features that participants used are the speed, the direction of the movement, and the change in the shape and size of the bounding boxes.

\section{Conclusion}
In this paper, we presented Transformer-based models for predicting pedestrians' crossing action decision in front of vehicles from an egocentric perspective. We have shown that using a simple and lightweight type of inputs (i.e., bounding boxes) with Transformers networks achieves high performance and outperforms the state-of-the-art models. Our findings clearly show the benefits when jointly training action and trajectory for pedestrian crossing action prediction. Furthermore, We introduced the CP2A simulated dataset which confirms our results on the PIE dataset with an accuracy and F1-score of 0.9. In addition, we have improved our model performance by applying Transfer learning from the CP2A dataset to the PIE dataset. Finally, we conducted the ``human attention to bounding boxes'' experiment that confirmed our first hypothesis and showed that humans can also predict the future by merely giving attention to the bounding boxes.

\section*{Acknowledgement}
This work was carried out in the framework of the OpenLab ``Artificial Intelligence'' in the context of a partnership between INRIA institute and Stellantis company.

\end{document}